# Comparative Analysis of Machine Learning and Deep Learning Algorithms for Detection of Online Hate Speech

Tashvik Dhamija[1], Anjum[2], Rahul Katarya[3]

[1,2,3] Delhi Technological University, New Delhi, India-110042
tashvik369@gmail.com, anjum_2792@yahoo.com, rahuldtu@gmail.com

**Abstract.** In the day and age of social media, users have become prone to online hate speech. Several attempts have been made to classify hate speech using machine learning but the state-of-the-art models are not robust enough for practical applications. This is attributed to the use of primitive NLP feature engineering techniques. In this paper, we explored various feature engineering techniques ranging from different embeddings to conventional NLP algorithms. We also experimented with combinations of different features. From our experimentation, we realized that roBERTa (robustly optimized BERT approach) based sentence embeddings classified using decision trees gives the best results of 0.9998 F1 score. In our paper, we concluded that BERT based embeddings give the most useful features for this problem and have the capacity to be made into a practical robust model.

## 1 Introduction

Hate speech is when an individual or a group of individuals attack or use pejorative or discriminatory language against a community based on traits like origin, sexuality, ethnicity, religious background, socioeconomic strata, race, gender and various other attributes. When such activity exists on social media portals, blogs, creative content and other media available on the internet, it is said to be Online Hate Speech (OHS)[1], [2]. Social media giants have seen exponential growth rates of users engaging with their platforms with an increase of accessibility of quality internet and availability of affordable devices in the remotest parts of the world. Twitter reported a massive 34% increase in users in the second quarter of 2020[3]. About 6000 tweets per second get posted on Twitter[4]. It showcases the intensity of these platforms exposing everyone to post, share and tweet whatever one feels. It brings about a need to keep a check on the content posted online. Often, users use these platforms to propagate hate and aggression against communities and individuals. They tend to use abusive language and target personality traits of the target community such as gender, socio-political status to inflict rage and community hatred against them. All of this is said to be OHS. It is toxic for not only the victims but also the social community as a whole.
They use such content to attain political, social or economic powers or superiority like generating a vote bank for parties, influencing decisions in favour of the majority. Many times, the offender uses it to condescend or belittle marginalized or minority communities. The offender can gain support from people with similar opinions and can stir propaganda or non-propaganda-based fights to scare or threaten communities and show off their assumed "superiority". Another mission is to be able to influence young minds by instilling biases and extreme opinions. Moreover, they are prone to an almost constant stream of information which they may not have the critical skills to filter and navigate. Hence, even without revealing their identity, a perpetrator can influence children through direct and indirect means. All of this boils down to a high requirement of a full proof system that can detect OHS and can help these social media platforms to get rid of such content to maintain harmony. We have seen Twitter, Facebook and Instagram take various measures to do so. Whereas, the current methods require individually judging data and are prone to bias whilst not being very efficient as they are based on regulations set by the companies itself[5].

In the paper, we compare different machine and deep learning models on a dataset of tweets using various feature engineering techniques and try to determine which algorithm and feature engineering method is the most appropriate for this problem statement.

## 2 Literature Survey

The literature survey is bifurcated into three broad classes:
   a) Based on feature engineering
   b) Based on an algorithm for classification
   c) Based on hate speech for other languages than just English

### 2.1 Based on feature engineering

This survey[6] focuses on the different techniques used for feature engineering of a dataset containing hate speech and non-hate speech text such. It talked about frequency-based techniques, sentiment analysis and word embeddings-based techniques. Then, it discusses the different Machine and Deep Learning-based algorithms used by different authors in their papers alongside the used feature engineering techniques. The current methods used in the task of detection of hate speech using Machine or Deep learning algorithms make use of manually handcrafted features that occupy a large feature space. This paper experiments[7] with automatically selected features and compares its performance with manual feature engineering without feature selection. The automatic selection of features is based on calculating the importance of each feature using Logistic Regression and discarding all features having importance below the selected threshold. The paper explains[8] a new algorithm to identify hate speech messages. Unlike the traditional semantic and syntactic approaches of feature engineering, the algorithm is solely implemented on feed's metadata using the Random Forest algorithm for classification. Their analysis indicates that metadata associated with interaction and structure of tweets has high relevance in identifying the content. Whereas, the metadata of the Twitter accounts has less use in the classification process. Another technique of feature engineering was explored which uses unigrams and patterns in texts to create the feature space. They also explore how these features improve results when used along with other sentiment and semantic-based features[9]. Another attempt to generate new distributed low dimension features that improve accuracy was done where the authors [10] train their embeddings called Comment Embeddings. This uses paragraph2vec and BoW to generate distributed low-level representations that are used to classify hate speech. This research study explains [11] a novel transfer learning algorithm based on the pre-trained model BERT. It evaluates if the embeddings generated by the model can be fine-tuned to so that they can capture the hateful context in them. This is seen as a solution to the problem of lack of labelled data or inability to improve generalization property. The process of fine-tuning is carried out by various methods.

### 2.2 Based on an algorithm for classification

This article gives[12] a dataset that is created using 2285 definitions from the Urban dictionary platform that are classified as misogynistic and non-misogynistic by three annotators of knowledge of the domain. The authors of this paper[13] compare different machine and deep learning models on 25K Twitter dataset[14]. They utilize different feature engineering models to determine which ML or DL algorithm is most appropriate for this problem. Taking a different perspective of the problem statement, authors of this paper[15] use word embeddings with machine and deep learning models to identify communities vulnerable to hate speech after classifying text as hate speech. For detection of abusive content in posts and comments online, this paper [16]creates a dataset from Yahoo News and Finance portals and experiments with Vowpal Wabbits's Regression for various types of feature spaces. To take the problem a step further, the authors [17] not only classifies Hate Speech in a binary fashion but further classifies it into subdivisions like Sexism, Racism etc. They use basic frequency and dictionary-based features to do so. An alternative approach taken by the authors of this paper[14] is to classify data into three classes, separating offensive from hate. They create a, now widely used, dataset from Twitter to accomplish their task.

### 2.3 Based on hate speech for other languages

The authors of this paper[18] use linguistic features and various machine learning algorithms to classify hate speech, not only in English but also in Spanish and to understand what features and models are good for the same. Similarly, this paper[19] uses two lists to translate Hinglish to English, followed by classification using LSTM based deep learning model to classify hate speech in Hinglish. Another attempt to expand this problem beyond English was made to the Arabic language where these authors compare a variety of machine and deep learning algorithms to classify data that is directly translated to English from Arabic. They used a self-created dataset containing balanced texts for hate and non-hate[20].

## 3 Experiments

### 3.1 Dataset
The dataset used in our experiments is a combination of the following two datasets:

a) 25k Twitter dataset[14]: This dataset comprises of 24,802 labelled tweets which were randomly sampled out of 85.4 million tweets and were labelled into hate, offensive and neither. We make this dataset binary by considering offensive and neither as non-hate.
b) Hate speech and personal attack dataset in English social media[21]: This a dataset available on zenodo.org. It is a binary dataset collected for the European project for countering hate speech.

### 3.2 Data Cleaning
NLTK library in python is used to perform the following data cleaning techniques to get a dataset that can be used to feature engineer and then feed into algorithms for training.

a) **Regex:**
Regular expressions are used to remove Twitter profile tags, hashtags, URLs and numbers. This output each tweet in for of a sequence of words only.
b) **Stop Words Removal:**
Stop words are words that are very commonly found in a sequence of texts and repeat very often. These words don't particularly add any meaning or value to the sentence. Removal of stop words improves accuracy as it helps in better feature engineering, decreases the length of sequences and removes words that don't hold much value yet repeat a lot.
c) **Lemmatizing:**
A word can exist in different tenses and participles in the English language. We use lemmatizing to replace all different forms of the same word by replacing it with one common word so that all different forms aren't treated as different words.

### 3.3 Feature Engineering
Feature engineering is the most important part of Machine learning to make any dataset usable for training any machine learning or deep learning model. There exist various types of features like semantic, lexographic, sentiment-based and word embeddings. It is very important to decide what features we use to optimize the accuracy of our models. The following are the feature engineering techniques used in this paper.

a) **TF-IDF:**
Term Frequency-Inverse Document Frequency (TF-IDF) is a technique used to give weighted importance of a word or a phrase in a document or a corpus. We used it to create feature vectors using the most common n-grams in our dataset[22].
b) **Bag of Words (BoW):**
BoW is a technique used to numerically express a document in our corpus using the number of times a specific word or n-gram occurs[6].

c) **Sentiment Analysis:**
Sentiment Analysis is the judgement of the emotion displayed in a document or sentence. It is numerically measured in a range of -1 to 1, 1 being positive, -1 being negative and 0 being neutral[23].

d) **Document Embeddings (Doc2Vec):**
Doc2Vec algorithm is a technique used to generate sentence embeddings regardless of the length of the sentence. It is an extension to the famous word2vec [24] model used to learn word embeddings except another matrix called the paragraph ID is also trained alongside the word embeddings which in turn, forms the sentence embeddings. This is used to generate a feature vector to numerically represent our dataset.

e) **Sentence Embeddings (BERT):**
Bidirectional Encoder Representations from Transformers (BERT) [25] is a transformer that uses the attention mechanism to generate encodings of a sequence of words. This was developed by Google AI Language. Pretrained RoBERTa model is used in this paper to generated sentence embeddings. We will be using RoBERTa [26], an optimized BERT model produced by Facebook.

f) **GloVe Word Embeddings:**
Global Vectors (GloVe) [27] is an algorithm designed to generate word embeddings. In this paper, it is used to convert text data into numerical form to make it usable for RNN based models that will be explained later in the paper.

g) **Word2vec Embeddings:**
Word2vec [24] is an algorithm that uses a neural network to learn word embeddings. Its goal is to estimate each word's position in a multi dimension vector space based on the similarity of different words.

### 3.4 Classifier Models

Once the feature engineering algorithm is used to convert the corpus of tweets into a usable dataset to train models, it is important to identify and select which algorithm is used to conduct the binary classification. We use the following algorithms for the same.

a) **Logistic Regression (LR)**
This a machine learning model that uses the sigmoid function to learn and predict the probability of the input belonging to a specific class [28]. This probability is compared to a threshold to attain the class of the input. We use the binary form of this algorithm.

b) **Decision Tree (DT)**
This is a machine learning algorithm with a tree-like structure that contains nodes at different levels [29]. This structure is learnt such that, according to the input, the class can be determined.

c) **Random Forest (RF)**
This is an ensemble-based machine learning algorithm wherein it consists of a large number of uncorrelated decision trees. Each tree gives a prediction and the class with most votes stands as the output of the random forest [30].

d) **Naïve Bayes (NB)**
   This is a machine learning algorithm based on the Bayes Theorem wherein, it assumes that each feature is an independent variable and uses the mathematical form of the theorem to give an output [31].
e) **Recurrent Neural Networks (RNN)**
   This is a class of neural networks that takes into consideration the state due to the previous input as well as the current input to produce the output. It is a Deep Learning algorithm and hence is data thirsty. We will use two types of RNNs, i.e., Long short-term memory (LSTM) [32] and Gated Recurrent Units (GRU) [33].

# 4 Results

We have experimented with different combinations of feature engineering with different machine learning algorithms. Also, we have used Deep Learning-based models with different word embedding algorithms. The dataset is broken into training and testing dataset with the ratio 70:30. The following are the accuracies that we obtained.

## 4.1 Using Bag of Words with n-grams and Sentiment Analysis

In the first experiment shown in table 1, we used Count Vectorizer along with n-grams. The grams are formed across the n-gram range of n=1 to n=6. The most common 1000 features were taken. Sentiment analysis was also done using TextBlob[23] and taken as the 1001st feature. The formed 1001 features were used with Logistic Regression, Random Forest, Decision Tree and Naive Bayes.

TABLE 1: CLASSIFICATION USING BOW AND SENTIMENT ANALYSIS

|    | Accuracy | Precision | Recall | F1 Score |
|----|----------|-----------|--------|----------|
| LR | 0.9242   | 0.9287    | 0.9018 | 0.9151   |
| DT | 0.9063   | 0.9223    | 0.8719 | 0.8964   |
| RF | 0.9205   | 0.9334    | 0.8910 | 0.9117   |
| NB | 0.8194   | 0.9027    | 0.7422 | 0.8146   |

## 4.2 Using Tf-Idf with n-grams and Sentiment Analysis

In the second experiment, we used TF-IDF Vectorizer along with n-grams. The grams are formed across the n-gram range of n=1 to n=6. Most common 1000 features were taken. Sentiment analysis was also done using TextBlob and taken as the 1001st feature. The formed 1001 features were used with Logistic Regression, Random Forest, Decision Tree and Naive Bayes the results are shown in table 2.

TABLE 2: CLASSIFICATION USING TF-IDF AND SENTIMENT ANALYSIS

|    | Accuracy | Precision | Recall | F1 Score |
|----|----------|-----------|--------|----------|
| LR | 0.9253   | 0.9187    | 0.9119 | 0.9153   |
| DT | 0.9074   | 0.9242    | 0.8727 | 0.8977   |
| RF | 0.9201   | 0.9074    | 0.9105 | 0.9090   |
| NB | 0.8615   | 0.8822    | 0.8174 | 0.8485   |

### 4.3 Using Doc2Vec

In the third experiment, we used the Doc2Vec algorithm using the gensim library to obtain 500 document embeddings for each sentence. The formed 500 features were used with Logistic Regression, Random Forest, Decision Tree and Naive Bayes. The results are presented in table 3.

TABLE 3: CLASSIFICATION USING DOC2VEC

|    | Accuracy | Precision | Recall | F1 Score |
|----|----------|-----------|--------|----------|
| LR | 0.9151   | 0.8981    | 0.9079 | 0.9030   |
| DT | 0.9062   | 0.9223    | 0.8719 | 0.8964   |
| RF | 0.9140   | 0.8920    | 0.9105 | 0.9012   |
| NB | 0.9032   | 0.9062    | 0.8776 | 0.8917   |

### 4.4 Using Sent2Vec and Sentiment Analysis

In the fourth experiment, we used the Sent2Vec algorithm using the "roberta-large-nli-stsb-mean-tokens" model based on BERT universal sentence encoder[25], [26] to obtain 1024 feature embeddings for each sentence shown in table 4. Sentiment analysis was also done using TextBlob and taken as the 1025th feature. The formed 1025 features were used with Logistic Regression, Random Forest, Decision Tree and Naive Bayes.

TABLE 4: CLASSIFICATION USING SENT2VEC AND SENTIMENT ANALYSIS

|    | Accuracy | Precision | Recall | F1 Score |
|----|----------|-----------|--------|----------|
| LR | 0.9247   | 0.9007    | 0.9260 | 0.9132   |
| DT | 0.9998   | 1.0       | 0.9996 | 0.9998   |
| RF | 0.9683   | 0.9391    | 0.9881 | 0.9630   |
| NB | 0.8427   | 0.8338    | 0.8132 | 0.8234   |

### 4.5 Using TF-IDF with n-grams along with Doc2Vec and Sentiment analysis

In the fifth experiment, we combined three different techniques of feature engineering in NLP. We used the top 1000 n-grams vectorized using TF-IDF vectorizer. We also generated 500 feature embeddings for each sentence using gensim's Doc2Vec algorithm. Finally, we performed Sentiment analysis using TextBlob. We concatenated all these features to form a 1501 feature vector. The formed 1501 features were used with Logistic Regression, Random Forest, Decision Tree and Naive Bayes. The results are presented in table 5.

TABLE 5: CLASSIFICATION USING DOC2VEC, TF-IDF AND SENTIMENT ANALYSIS

|    | Accuracy | Precision | Recall | F1 Score |
|----|----------|-----------|--------|----------|
| LR | 0.9567   | 0.9402    | 0.9604 | 0.9502   |
| DT | 0.9481   | 0.9178    | 0.9624 | 0.9396   |
| RF | 0.9530   | 0.9238    | 0.9680 | 0.9532   |
| NB | 0.9419   | 0.9243    | 0.9425 | 0.9333   |

### 4.6 Using BoW with n-grams along with Doc2Vec and Sentiment analysis

In the sixth experiment shown in table 6, we combined three different techniques of feature engineering in NLP. We used the top 1000 n-grams vectorized using BoW vectorizer. We also generated 500 feature embeddings for each sentence using gensim's Doc2Vec algorithm. Finally, we performed Sentiment analysis using TextBlob. We concatenated all these features to form a 1501 feature vector. The formed 1501 features were used with Logistic Regression, Random Forest, Decision Tree and Naive Bayes.

TABLE 6: CLASSIFICATION USING DOC2VEC, BOW AND SENTIMENT ANALYSIS

|    | Accuracy | Precision | Recall | F1 Score |
|----|----------|-----------|--------|----------|
| LR | 0.9564   | 0.9378    | 0.9621 | 0.9498   |
| DT | 0.9502   | 0.9307    | 0.9549 | 0.9426   |
| RF | 0.9531   | 0.9231    | 0.9687 | 0.9453   |
| NB | 0.9445   | 0.9282    | 0.9447 | 0.9363   |

### 4.7 Using Deep Learning Model with word-embeddings

In the seventh experiment presented in table 7, we have used GloVe pretrained embeddings. We also trained our embeddings using Gensim's Word2vec[24] algorithm using our dataset as corpus. Both these embeddings are used to train a LSTM and GRU model each.

TABLE 7: CLASSIFICATION USING DEEP LEARNING MODELS WITH WORD EMBEDDINGS

|         | Embeddings Used | Accuracy | Precision | Recall | F1 Score |
|---------|-----------------|----------|-----------|--------|----------|
| Bi-LSTM | GloVe           | 0.9291   | 0.9062    | 0.8992 | 0.9014   |
| Bi-GRU  | GloVe           | 0.9320   | 0.8881    | 0.9195 | 0.9022   |
| Bi-LSTM | Word2vec        | 0.9222   | 0.9009    | 0.8978 | 0.8972   |
| Bi-GRU  | Word2vec        | 0.9209   | 0.8983    | 0.9004 | 0.8966   |

## 5 Conclusion and Future Work

In this paper, we have compared different machine learning algorithms and deep learning architectures trained on the same dataset to classify Online hate speech. Moreover, we have compared different feature engineering techniques and their combinations to see what type of features are the most appropriate for this problem.

Through our experiments, we have concluded that TF-IDF and BoW algorithms have similar accuracies, with TF-IDF being a little better in most cases. Embeddings generated feature vectors perform better than conventional NLP features. Doc2Vec, when combined with n-grams features using TF-IDF or BoW perform even better than Doc2Vec used along. The best performance was showcased by state-of-the-art BERT sentence embeddings with almost perfect accuracies using a Decision Tree algorithm.

Deep Learning models conventionally perform very well given that they get huge datasets. This can be seen as RoBERTa embeddings that are trained on millions of articles and is able to give us robust embeddings which are used to get almost 100% accuracy using a decision tree.

Future work can be done by expanding the variety of sequences of words in the dataset by increasing the number of samples available and by sampling different social media networks to make the classifiers more robust and accurate. Various other combinations of feature engineering techniques can be compared and all embedding generating algorithms can be trained from scratch rather than using pre-trained models. Another direction to work is to experiment with multimodal data such as images and speech. Models can be used to identify hate speech in pictures and audios on the internet. The gaps in this field of research are majorly based on the missing perception of practicality. The classifiers should be trained such that false positives and negatives shall be low to protect the reputation of the user. Moreover, consideration of the fact that hate is not binary but rather of various types such as racism, sexism and more is lacking. These are gaps in this field of research that have potential for future work. All these future experiments will push towards models that have high accuracies and help make the internet a safer place for individuals to virtually exist in.